\documentclass[conference]{IEEEtran}
\IEEEoverridecommandlockouts
\usepackage{cite}
\usepackage{amsmath,amssymb,amsfonts}
\usepackage{algorithmic}
\usepackage{graphicx}
\usepackage{textcomp}
\usepackage{xcolor}
\usepackage{caption}
\usepackage{subcaption}
\usepackage{hyperref}
\hypersetup{
    colorlinks=true,
    citecolor=blue,
    linkcolor=blue,
    filecolor=black,
    urlcolor=cyan
}
\usepackage[printonlyused,withpage]{acronym}
\newacro{ann}[ANN]{artificial neural networks}
\newacro{eca}[ECA]{emergency collision avoidance}
\newacro{afs}[AFS]{active front steering}
\newacro{abs}[ABS]{anti-lock braking system}
\newacro{dyc}[DYC]{direct yaw moment control}
\newacro{tcs}[TCS]{traction control system}
\newacro{lstm}[LSTM]{long-short term memory}
\newacro{rnn}[RNN]{recurrent neural network}
\newacro{cnn}[CNN]{convolutional neural networks}
\newacro{roi}[ROI]{region of interest}
\newacro{pca}[PCA]{principle component analysis}
\newacro{adas}[ADAS]{advanced driver-assistance systems}
\newacro{ans}[ANS]{autonomous navigation systems}
\newacro{ass}[ASS]{active safety systems}
\newacro{ebm}[EBMs]{energy based models}
\newacro{ssl}[SSL]{self-supervised learning}
\newacro{dnn}[DNN]{deep neural networks}
\newacro{nce}[NCE]{noise contrastive estimation}
\newacro{sm}[SM]{score matching}
\newacro{mcmc}[MCMC]{Markov chain Monte Carlo}
\newacro{bnn}[BNN]{Bayesian neural network}
\newacro{nir}[NIR]{near-infrared}
\newacro{cnn}[CNN]{convolutional neural networks}
\newacro{mle}[MLE]{maximum likelihood estimate}
\newacro{bnn}[BNN]{Bayesian neural networks}
\newacro{nlp}[NLP]{natural language processing}
\newacro{cv}[CV]{computer vision}
\newacro{mcd}[MC-Dropout]{Monte Carlo dropout}
\newacro{sgld}[SGLD]{Stochastic Gradient Langevin Dynamic}
\newacro{sgd}[SGD]{stochastic gradient descent}
\newacro{cd}[CD]{contrastive divergence}
\newacro{vo}[VO]{visual odemetry}
\newacro{em}[EM]{electronic motor}
\newacro{tos}[TOS]{transmission output shaft}
\newacro{fdr}[FDR]{final drive ratio}
\newacro{imu}[IMU]{inertial measurement unit}
\newacro{vae}[VAE]{variational auto-encoder}
\newacro{byol}[BYOL]{Bootstrap your own latent}
\newacro{ins}[INS]{Inertial Navigation System}
\newacro{hlvm}[HLVM]{Hierarchical Latent Variable Model}
\newacro{cvae}[CVAE]{Conditional Variational Autoencoder}
\newacro{elbo}[ELBO]{Evidence Lower Bound}
\newacro{hcvae}[HCVAE]{Hierarchical Conditional Variational Autoencoder}
\newacro{mae}[MAE]{Masked Autoencoder}
\newacro{vit}[ViT]{Vision Transformer}
\newacro{cvt}[CvT]{Convolutional vision Transformer}
\newacro{gp}[GP]{Gaussian Process}

\newacro{fps}[FPS]{frames per second}
\newacro{svm}[SVM]{Support Vector Machines}
\newacro{knn}[KNN]{K-nearest Neighbors}
\newacro{da}[DA]{Discriminant Analysis}
\newacro{pls}[PLS]{Partial Least Squares}
\newacro{rmse}[RMSE]{Root-mean-square Error}

\def\BibTeX{{\rm B\kern-.05em{\sc i\kern-.025em b}\kern-.08em
    T\kern-.1667em\lower.7ex\hbox{E}\kern-.125emX}}
\begin{document}

\title{Estimating friction coefficient using generative modelling}

% \author{
% \IEEEauthorblockN{Mohammad Otoofi\textsuperscript{*} \quad William J.B. Midgley \quad Leo Laine \quad Henderson Leon \quad Laura Justham \quad James Fleming}
% \thanks{\textsuperscript{*}Corresponding Author: M.Otoofi@lboro.ac.uk}
% }

\author{\IEEEauthorblockN{Mohammad Otoofi}
\IEEEauthorblockA{\textit{Wolfson School of Engineering} \\
\textit{Loughborough University}\\
Loughborough, UK \\
M.Otoofi@lboro.ac.uk}
\and
\IEEEauthorblockN{William J.B. Midgley}
\IEEEauthorblockA{\textit{Institution of Mechanical Engineers} \\
\textit{University of New South Wales}\\
Sydney, Australia \\
W.Midgley@unsw.edu.au}
\and
\IEEEauthorblockN{Leo Laine}
\IEEEauthorblockA{\textit{Volvo Group Truck Technology} \\
Gothenburg, Sweden \\
Leo.Laine@volvo.com}
\and
\IEEEauthorblockN{Henderson Leon}
\IEEEauthorblockA{{\textit{Volvo Group Truck Technology}}\\
Gothenburg, Sweden \\
Leon.Henderson@volvo.com}
\and
\IEEEauthorblockN{Laura Justham}
\IEEEauthorblockA{\textit{Wolfson School of Engineering} \\
\textit{Loughborough University}\\
Loughborough, UK \\
L.Justham@lboro.ac.uk}
\and
\IEEEauthorblockN{James Fleming}
\IEEEauthorblockA{\textit{Wolfson School of Engineering} \\
\textit{Loughborough University}\\
Loughborough, UK \\
J.Fleming@lboro.ac.uk}
}

\maketitle

\begin{abstract}

It is common to utilise dynamic models to measure the tyre-road friction in real-time. Alternatively, predictive approaches estimate the tyre-road friction by identifying the environmental factors affecting it. This work aims to formulate the problem of friction estimation as a visual perceptual learning task. The problem is broken down into detecting surface characteristics by applying semantic segmentation and using the extracted features to predict the frictional force. This work for the first time formulates the friction estimation problem as a regression from the latent space of a semantic segmentation model. The preliminary results indicate that this approach can estimate frictional force.

\end{abstract}


\section{Introduction}

It is said that one of the critical factors in available frictional force is surface condition \cite{ghandour2010tire,roychowdhury2018machine}. Surface condition can provide valuable information about its  anti-sliding  performance \cite{pu2020road}. At the moment, dynamic models are mostly utilized to calculate the current frictional force \cite{ruvzinskas2017magic, zhao2017estimation, chen2016novel}. Alternatively, predictive approaches can estimate frictional force by analyzing carefully the environmental factors, e.g. surface condition and weather, and proactively respond to hazardous situations.

A variety of predictive solutions have been proposed by either utilizing expensive sensors, such as expensive \ac{nir} cameras \cite{jonsson2014road}, or by conducting experiments under laboratory conditions, e.g. using a microphone mounted in the tyre \cite{masino2017road}. Many other similar techniques have been suggested in the literature \cite{khaleghian2017technical,jonsson2014road,masino2017road,vsabanovivc2020identification, li2015comprehensive, ruvzinskas2017magic,khaleghian2017application} but unfortunately many of these are not scalable or practical in real-world scenarios. This work aims to propose a scalable and deployable solution using a simple dashcam and machine learning algorithms. Using semantic segmentation algorithm, it is aimed to identify and extract critical information from the elements that exist in the scene to derive an accurate estimation of frictional force.

However, relying on a simple dashcam camera introduces new challenges.  Environmental noise like sun glare, shadows, and poor visibility can fool semantic segmentation algorithms easily \cite{goodfellow2014explaining, pfeuffer2020robust}. For this, a type of generative model called a \ac{hlvm} is utilized \cite{murphy2022probabilistic}. Using a \ac{hlvm} has two fold benefits. First, it enables robust learning to address the noisy frames captured by the dashcam, e.g. sun glare, shadows, etc. Second, it can provide the degree of uncertainty about the predictions which is considered essential for safety-related measurements used to make vehicle motion decisions.

More specifically, this work utilizes a \ac{cvae} to parameterize a \ac{hlvm}. Then, the extracted features by the \ac{cvae} are employed to estimate frictional force. The CityScapes \cite{Cordts2015Cvprw} dataset is used for semantic segmentation part. And, thanks to facilities provided by Volvo Trucks, this is tested on a new visual dataset created for vehicle dynamic response. This dataset offers the frames captured by the dashcam synced with the dynamic signals obtained by vehicle based sensors. 

The contributions of this work can be summarized as follows:
\begin{enumerate}
  \item A visual dataset is created for vehicle dynamic response. This dataset is utilized to train and test a regression model to estimate frictional force.
  \item While using semantic segmentation to identify surface condition is not new and was considered in  \cite{liang2019winter}, this work builds upon \cite{kohl2019hierarchical} by applying the \ac{cvae} generative model.
  \item Using a latent variable model like \ac{cvae} enables  regression to estimate the frictional force using the low-dimensional features extracted from the latent space of the \ac{cvae} model. This joint segmentation and friction regression has not previously been attempted, and has the advantage that the friction estimation can exploit the semantic segmentation model's understanding of surface types. The low-dimensional latent space used in the segmentation model has learned high level features or a broad internal representation of roads and environmental elements that can be used to estimate friction.
\end{enumerate}

% \begin{figure}[!h]
%      \centering
%          \includegraphics[width=0.4\textwidth]{img/noise.png}
%          \caption{Example of environmental noise \cite{pfeuffer2020robust}}
%          \label{fig:noise}
% \end{figure}

\section{Literature review}

Since applying semantic segmentation and regression jointly to estimate frictional force is a new idea, separate reviews are provided for semantic segmentation methods and the methods related to frictional force.

\subsection{Frictional force}

There are two sets of approaches in the literature to estimate the road-tyre friction coefficient: Developing accurate dynamic models or finding correlation between sensory data and friction-related parameters \cite{khaleghian2017technical}. Regarding dynamic model approaches, there are generally three types of dynamic models to derive tyre-road friction: wheel and vehicle dynamic models \cite{canudas2003new,doumiati2012vehicle}, tire model based \cite{pacejka1992magic,pacejka1997magic}, and slip-slope based approaches \cite{gustafsson1997slip,han2017adaptive}. These models rely on the vehicle's dynamic response which means to calculate the friction, the vehicle first needs to reach the part of the surface that the friction is intended to estimate. For instance, an ABS system measures the response of the vehicle to the environment and adjusts itself in real-time to the current conditions. An alternative solution is to use sensor-based solutions. This family of methods utilizes sensory equipment to measure the parameters affecting friction coefficient. Having identified the effective parameters, they can be used by statistical or learning methods to find correlation between those parameters and the friction coefficient. To identify these parameters, various types of sensor equipment have been employed in the literature from different types of cameras \cite{jonnarth2018camera} to acoustic sensors \cite{alonso2014board,kalliris2019machine}, and tire/road-related sensors \cite{yang2018convolutional,vsabanovivc2020identification}.

This work applies a predictive approach using a vision-based model to friction. The aim is to exploit the environmental information in a scene to estimate the normalised frictional force.

\subsection{Vision-based predictive model}

One common approach in the literature \cite{roychowdhury2018machine, rateke2019road, rateke2019road} using \ac{cnn} to estimate friction is to split the input images into small patches and use a classifier to categorize each patch to a surface type. Then, the detected surfaces are further classified into qualitative friction groups, e.g. high, medium, and low friction coefficient. \cite{vsabanovivc2020identification} applies the result of surface classification to determine Burckhardt tyre model parameters \cite{burckhardt1993fahrwerktechnik}.

Equipped with an Infrared camera, Jonsson, et.al. \cite{jonsson2014road} classified each pixel of the road image into dry,  wet,  icy  or  snowy using spectral analysis of the light reflected from surface. The collected data is classified by five machine learning methods, \ac{svm}, \ac{ann}, \ac{knn}, \ac{da}, and \ac{pls}. This method was developed for road weather stations and must be accompanied with a halogen illuminator.

\begin{figure}[!h]
     \centering
         \includegraphics[width=0.4\textwidth]{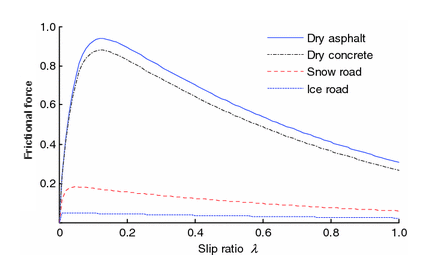}
         \caption{Friction coefficient graph \cite{zhao2017estimation}}
         \label{fig:friction_coe}
\end{figure}

The proposed method in this work is very similar to \cite{liang2019winter}. Unlike this previous work where the focus was on classifying the patches of input images, in this work semantic segmentation of the surface ahead uses a variant of U-Net architecture \cite{ronneberger2015u}.

\section{Methodology}

\subsection{Dataset}

As the approach adopted in this project is data-driven, one of the main concerns has been collecting the relevant data to the context of this problem. Thus, over the past two years, a lot of effort has been made to gather visual data for friction estimation. The proposed method consists of two steps: semantic segmentation and friction estimation. For the first step, semantic segmentation, there are publicly available datasets. In this work, the popular CityScapes dataset \cite{Cordts2015Cvprw} is utilized to provide images and labels for dry asphalt category. For the friction estimation step, the main concern is that there is no publicly available visual dataset for friction estimation. In this section, it is explained how the visual friction dataset used in this work was created. As the Cityscapes dataset only supports a `dry asphalt' category, here only the visual friction data collected for dry asphalt is going to be utilized. 

This section provides theoretical foundation of how friction coefficient is calculated using the vehicle dynamic responses to provide ground truth.

The graph presented in Fig \ref{fig:friction_coe} shows the relationship between frictional force, i.e. normalized force at road $F$, and slip ratio, $\lambda$, on different surfaces. The visual friction dataset consists of images of different surface types and conditions and their corresponding vehicle dynamic responses based upon which the slip ratio and frictional force can be derived.

To calculate the frictional force ground truth, the maximum utilised friction coefficient ($\mu$) is used. This can be calculated using maximum force point, $F_{max}$, and normal force ($N$), (\ref{eq:force_max}). Since the vehicle speed was low, the drag force was negligible and can be ignored.

\begin{equation}
F_{max} = \mu N\\
\Rightarrow \mu = \frac{F_{max}}{N}
\label{eq:force_max}
\end{equation}

By substituting Newton's second law $F = ma$ in \ref{eq:force_max}, the equivalent $\mu$ can be computed on a flat surface at low speed:

\begin{equation}
\mu = \frac{ma}{mg} = \frac{a}{g}
\label{eq:mu}
\end{equation}

Where $a$, $m$, and $g$ is the vehicle acceleration, vehicle mass, and gravitational constant, accordingly. Using (\ref{eq:mu}) enables calculating the average normalized frictional force over all wheels.

To achieve this, a GoPro\footnote{https://gopro.com/en/gb/shop/cameras/hero8-black/CHDHX-801-master.html} camera was used as a dashcam to capture footage. The dynamic responses of the vehicle were measured by an $RT3000$\footnote{https://www.oxts.com/products/rt3000-v3/}, which is an \ac{ins}, and vehicle measurement were taken form CAN BUSes ports available on the truck. Both $RT3000$ and CAN BUSes were connected to the Mlogger device. The Mlogger manages the incoming signals from different sources by creating a common timestamp. The data collected by the Mlogger contained $1700$ vehicle dynamic responses from which a few of them are required to calculate frictional force.

%All the test were carried out using a Volvo FH-$2430$ truck without trailer similar to that shown in Fig \ref{fig:truck}. 

% \begin{figure}[!h]
% \centering
% \begin{subfigure}[b]{.45\linewidth}
% \includegraphics[width=\linewidth]{img/camera_kit.jpeg}
% \caption{Camera and Toughbook setup}\label{fig:camera_kit}
% \end{subfigure}
% \begin{subfigure}[b]{.45\linewidth}
% \includegraphics[width=\linewidth]{img/fh.jpg}
% \caption{Test truck taken from Internet}\label{fig:truck}
% \end{subfigure}
% \caption{Data gathering module}
% \label{fig:datagatheringmodule}
% \end{figure}

\begin{figure}[!h]
     \centering
         \includegraphics[width=0.5\textwidth]{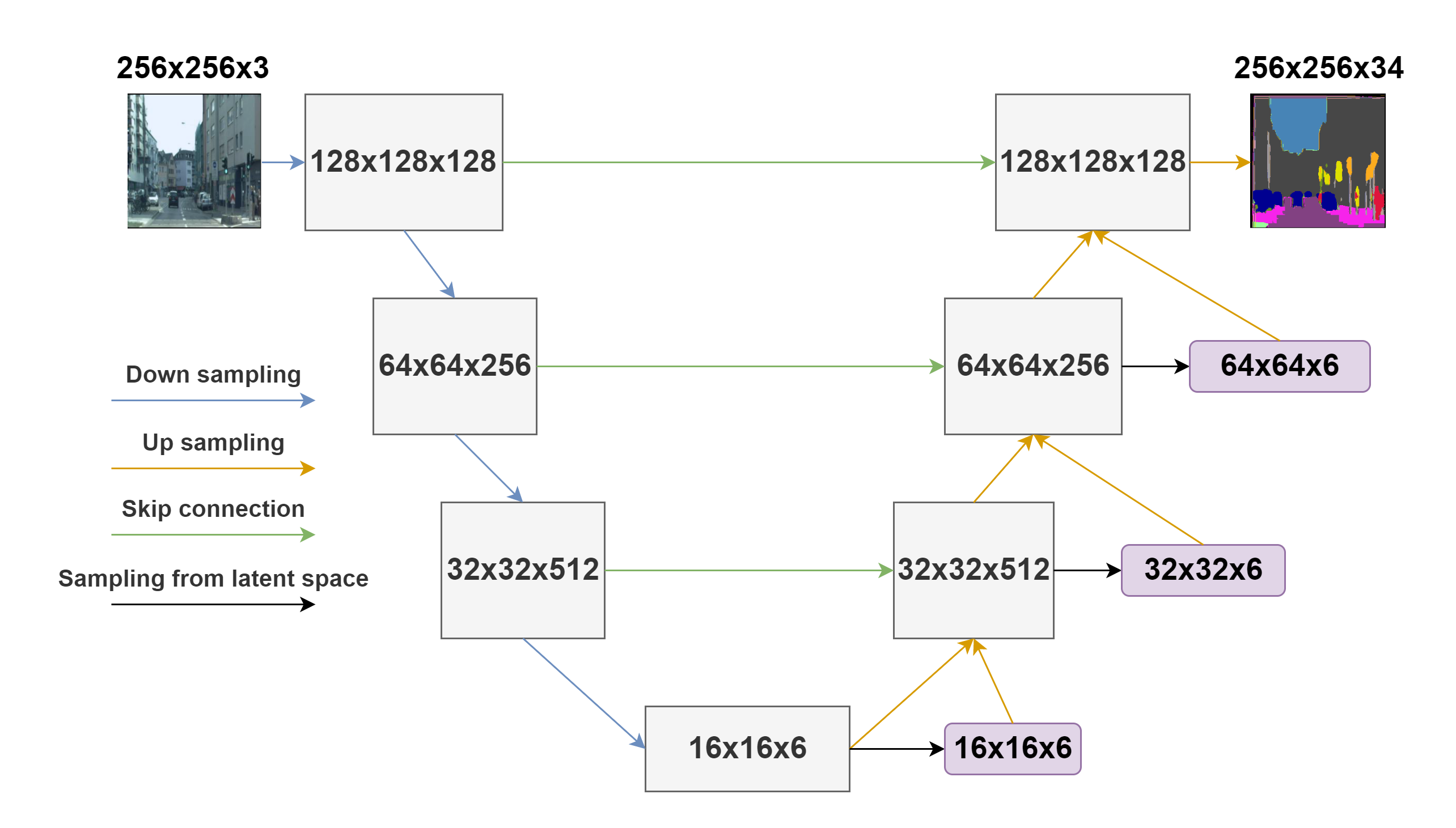}
        \caption{\ac{cvae} architecture trained on Cityscapes}
        \label{fig:SegArch}
\end{figure}

\subsection{\acl{cvae}}

Due to the lack of semantic understanding, discriminative modeling is unsuitable for safety-related applications. Alternatively, the conditional class distribution, $p(y|x)$, can be modeled by approximating joint distribution of input images ($x$) and segmentation masks ($y$), $p(x,y)$, using generative modeling. As computing this joint distribution is intractable, it is modeled by introducing a hierarchy of latent variables, $z$s, which are learned using variational inference. The hierarchical structure of latent variables improves the model's ability to learn the features that vary across space and scale. This can improve the generated segmentation masks by refining fine and coarse structures in predictions. The state-of-the-art method to parameterize the conditional distribution is \ac{cvae} \cite{sohn2015learning}. \ac{cvae} objective function is shown in Eq \ref{eq:lossHCVAE}.

\begin{equation}
\begin{aligned}
L_{HCVAE}(x,y) =&-\frac{1}{L}\sum_{1}^{L} E_{q_{z_{<i} \sim q_{\phi}(z|x,y)}}\\
&[KL(q_{i}(z_i|z_{<i},x,y)||p_{i}(z_i|x, z_{<i}))] \\
&+E_{q_{z} \sim q_{\phi} (z|x,y)}[\log p_{c} (y|x,z)]
\end{aligned}
\label{eq:lossHCVAE}
\end{equation}

Where $L$ is the number of latent variables and $KL$ denotes Kullback–Leibler divergence. The \ac{cvae} objective function is composed of multiple neural networks, such as recognition network $p_{c}(y|x,z)$, (conditional) prior network $p_{i}(z_i|x, z_{<i})$, and generation network $p_{\theta} (y|x,z)$. While in discriminative modeling the conditional distribution of $p(y|x)$ only generates the most likely answer due to its \ac{mle} nature, by introducing latent variables the \ac{cvae} enables modeling multiple modes in the conditional distribution, $p(y|x)$. Facing ambiguous, i.e. noisy, input images, \ac{cvae} can generate multiple hypotheses $(N)$ and thus, the uncertainty of the predictions can be computed by averaging the predicted probabilities obtained by sampling multiple times from the prior. In mathematical form, this is shown in (\ref{eq:answ}):

\begin{equation}
\begin{aligned}
p(y|x) &= \frac{1}{N}\sum_{z_{i}\sim q_{\phi}(z|x)}^{z_{N}} p_{c}(y|x,z_{i})
\end{aligned}
\label{eq:answ}
\end{equation}

\subsection{Friction estimation}

To estimate frictional force, a separate model from the semantic segmentation model is proposed. As there is no visual dataset that provides ground truth for both the segmentation tasks and frictional force estimation that would allow segmentation and friction estimation to be carried out by an end-to-end approach. Therefore, the frictional force estimation model must be trained separately on a different dataset. For a given input image, $x\in\mathbb{R}^{n\times m\times c}$, the segmentation model generates a segmentation mask, $y\in\mathbb{R}^{n\times m\times c^\prime}$ where $c^\prime$ is the number of categories in the dataset. Having semantically segmented an input image, the identified environmental elements are passed to the friction estimation module in the form of a one-hot vector, $\mathbf{s}\in\mathbb{R}^{n}$, e.g. $[0,0,1,0]$, along with the latent factor sampled from the latent space given the input, $\mathbf{z}\in\mathbb{R}^{L}$.

Here, the sampled latent variable encapsulates the elements that exist in the scene, e.g. weather, in order to provide contextual information. The sampled latent variable, $\mathbf{z}$, is utilized as the input to the friction model and the one-hot vector, $\mathbf{s}$, is concatenated to the mid-level features of the friction estimator as an additional feature. Since the friction coefficient is a continuous variable, $\mu\in[0,1]\subset\mathbb{R}$, it is needed to formulate the problem as regression where the aim is to map a set of features to a continues range, $g_{\phi}:h_{L+6}(R)\rightarrow \mu(R)$.

\subsection{Experiments}

\begin{figure}[h!]
\centering
\begin{subfigure}[b]{\linewidth}
\includegraphics[width=\linewidth]{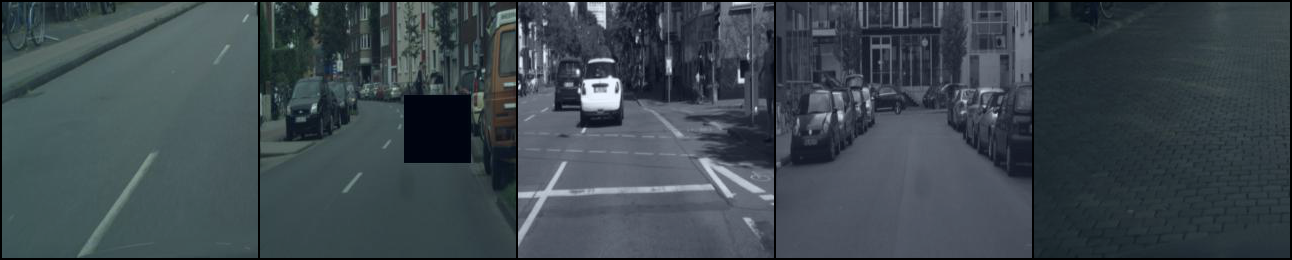}
\caption{Input images}
\label{fig:input}

\includegraphics[width=\linewidth]{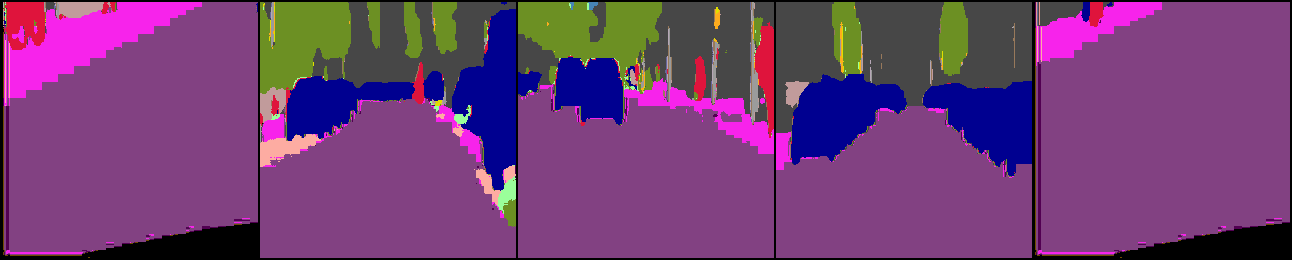}
\caption{Predicted segmentation masks}
\label{fig:predictions}

\includegraphics[width=\linewidth]{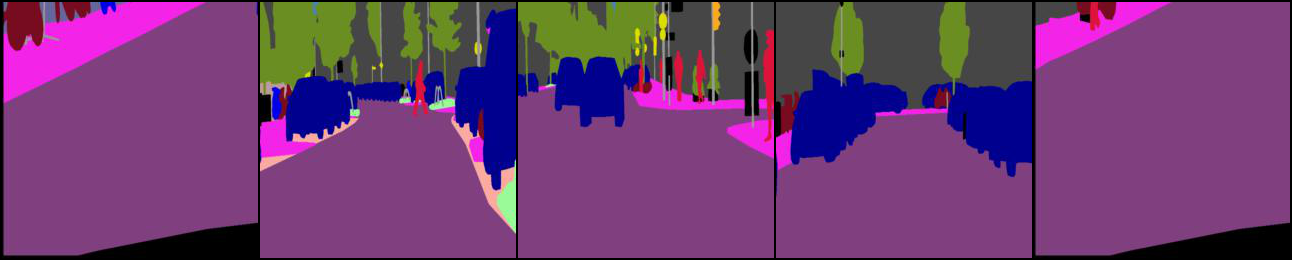}
\caption{Ground truth}
\label{fig:gt}

\end{subfigure}

\caption{\ac{cvae} prediction samples}
\label{fig:cvaeplots}
\end{figure}

Following the architecture proposed in \cite{kohl2019hierarchical},  a UNet architecture with three levels is implemented to parameterize the \ac{cvae}. Every level contain two residual layers and there are three latent spaces in total. The architecture is shown in Fig \ref{fig:SegArch}. As mentioned before, the Cityscapes dataset is used to train the segmentation model. The model is trained for about $1300$ iterations. In order to train a robust model, a number of augmentation strategies were applied to the training inputs. First, each image sample was cropped into 5 patches and then each patch was transformed by uniformly picking one of the following augmentation functions: color jittering, randomly erasing part of the image, blurring the image using Gaussian Blur, converting the image to gray scale, and randomly rotating the image between $0$ and $360$ degrees. The result of augmentation can be seen in Fig \ref{fig:input}. 

\begin{figure}[h!]
     \centering
         \includegraphics[width=0.4\textwidth]{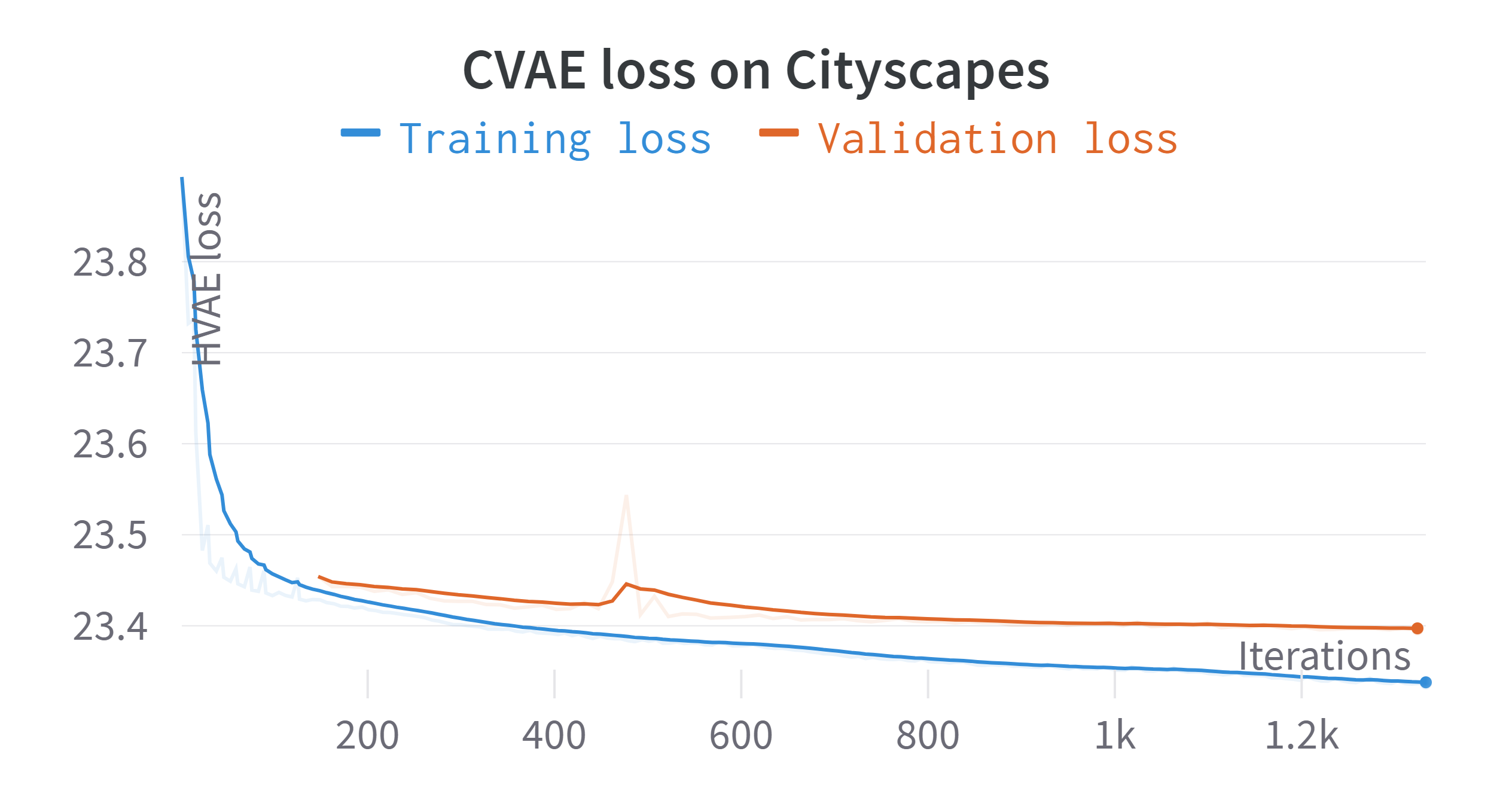}
        \caption{\ac{cvae} training curve for Cityscapes}
        \label{fig:finalloss}
\end{figure}

The hierarchical UNet is optimized using the objective function (\ref{eq:lossHCVAE}) similar to  \cite{kohl2019hierarchical}. The learning curves for this model are shown in Fig \ref{fig:finalloss}. Moreover, for qualitative assessment, Fig \ref{fig:cvaeplots} shows some of the predictions from the validation set. It is noteworthy to mention that  each of the predicted segmenting masks in Fig \ref{fig:predictions} is obtained by averaging $16$ predicted segmentation masks generated by sampling from the latent space $16$ times.

Having trained the semantic segmentation model, two experiments were carried out to estimate the normalized frictional force. In the first experiment, an end-to-end approach is applied where a deep \ac{cnn} is utilized to estimate the force. And in the second experiment, the semantic segmentation model is used as feature extractor to estimate the normalized frictional force shown in (\ref{eq:force_max}). In both experiments, the \ac{rmse} objective function is applied to train the models.

\begin{table}[]
\centering
\begin{tabular}{|c|c|}
\hline
\textbf{Method} & \textbf{IoU} \\ \hline
Ours       & $0.69$          \\ \hline
A Hierarchical Probabilistic U-Net \cite{kohl2019hierarchical}      & $0.62$          \\ \hline
\end{tabular}
\caption{IoU measured on Cityscapes validation set}
\label{Tab:iou}
\end{table}

\begin{figure}[!h]
     \centering
         \includegraphics[width=0.5\textwidth]{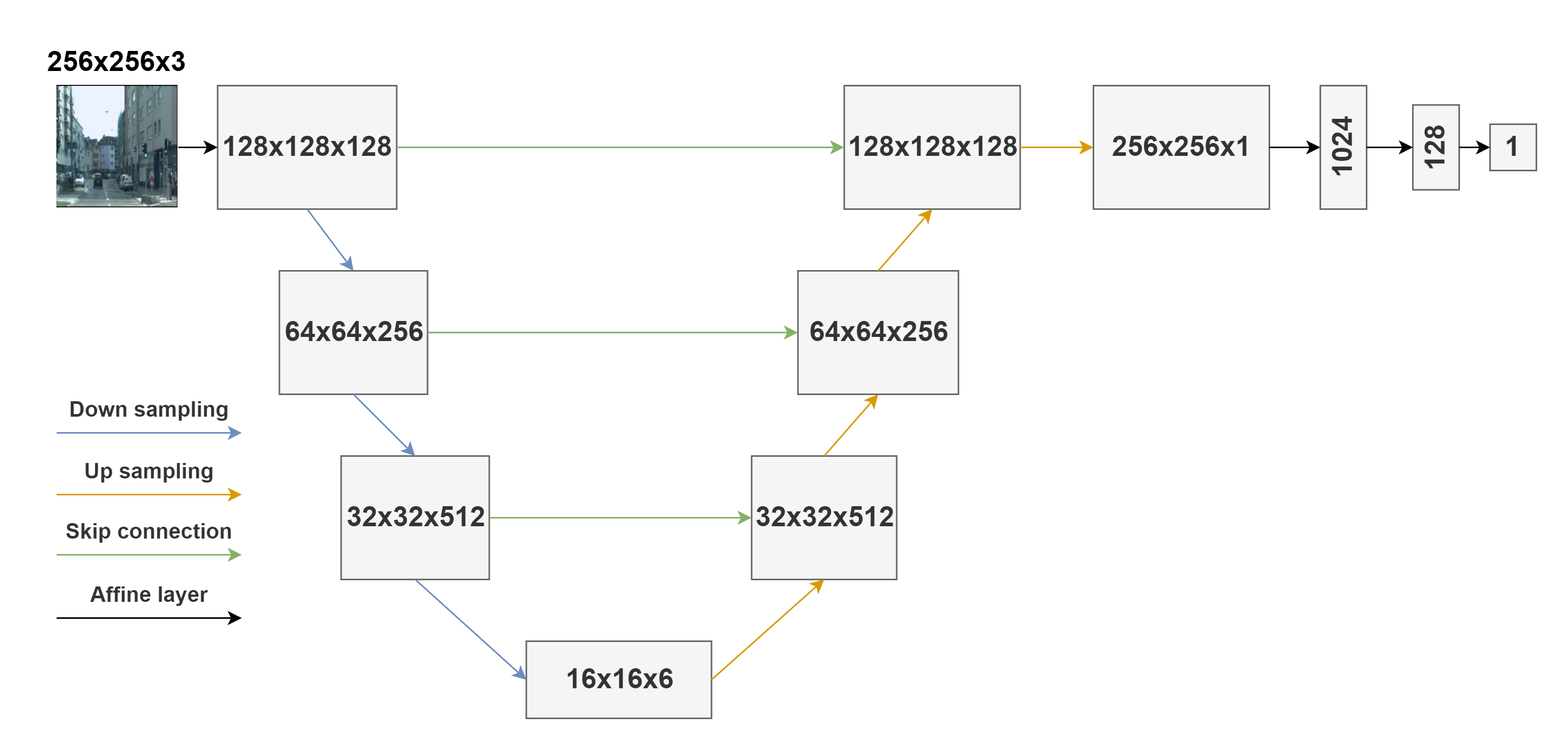}
        \caption{End-to-end friction estimator architecture}
        \label{fig:endtoendarch}
\end{figure}

For the first experiment, a UNet architecture similar to the segmentation model is adopted with this difference that it is a discriminative model and there is no latent spaces. Besides, after the last de-convolutional layer the spatial features are flattened and gone through three affine layers to estimate the force. The architecture is shown in Fig \ref{fig:endtoendarch}. This model is used as the baseline.

\begin{figure}[!h]
     \centering
         \includegraphics[width=0.5\textwidth]{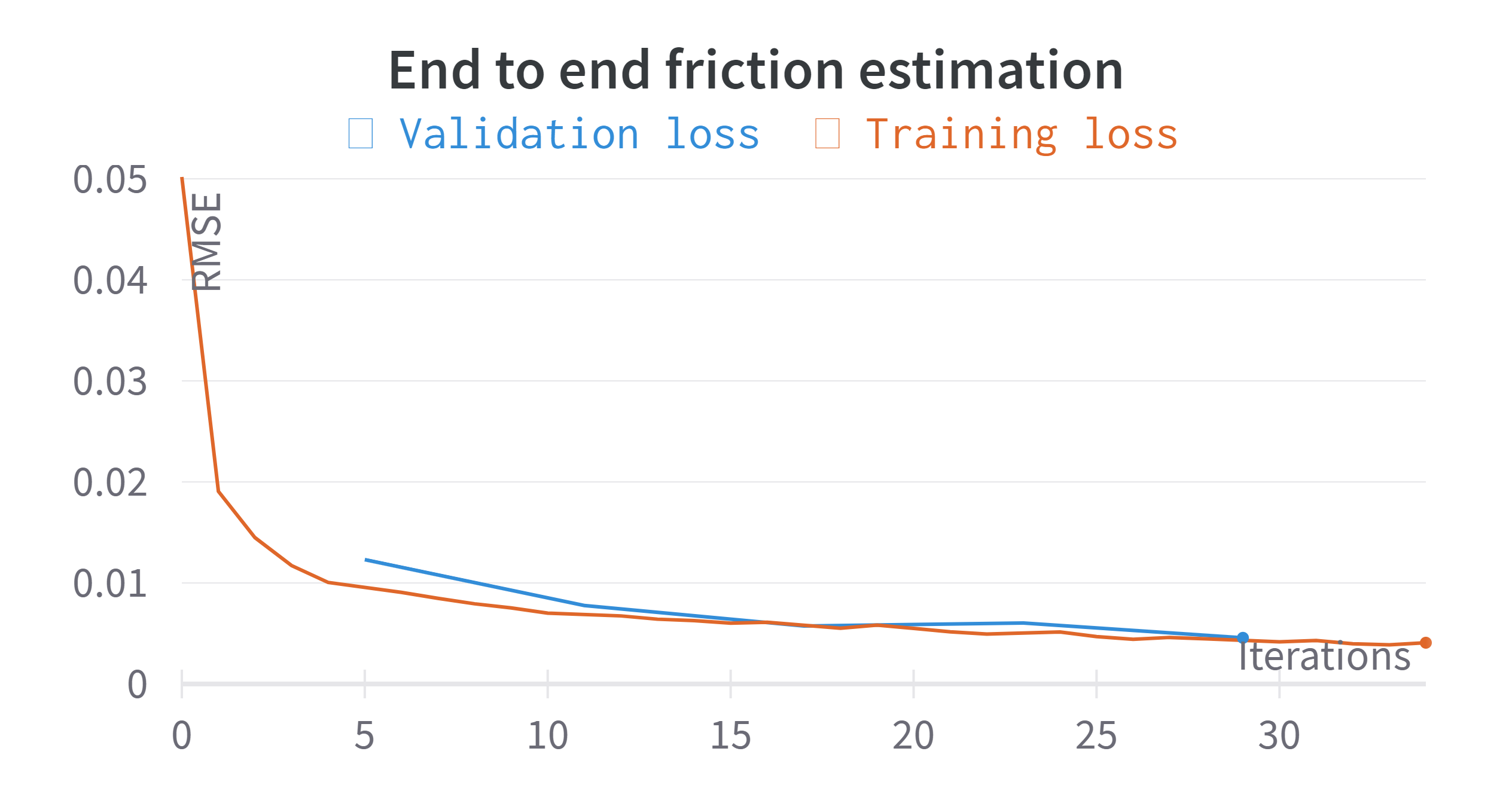}
        \caption{End-to-end model learning curve}
        \label{fig:endtoendlearningcurve}
\end{figure}

\begin{figure}[!h]
     \centering
         \includegraphics[width=0.5\textwidth]{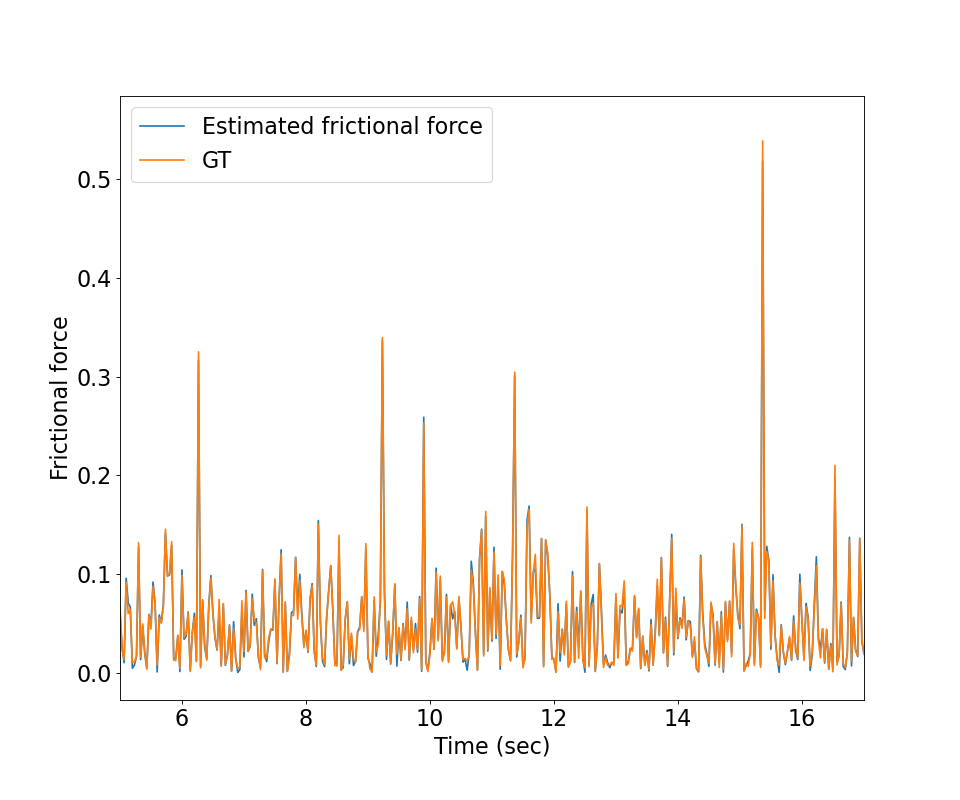}
        \caption{Estimated force against time}
        \label{fig:endToEndQualResu}
\end{figure}

The end-to-end model is trained for $35$ epochs and its learning curve can be seen in Fig \ref{fig:endtoendlearningcurve}. To demonstrate how the model performs on a validation set, a subset of consecutive frames from the validation set are selected and the estimated force is plotted against time. This plot is shown in Fig \ref{fig:endToEndQualResu}.

\begin{figure}[!h]
     \centering
         \includegraphics[width=0.5\textwidth]{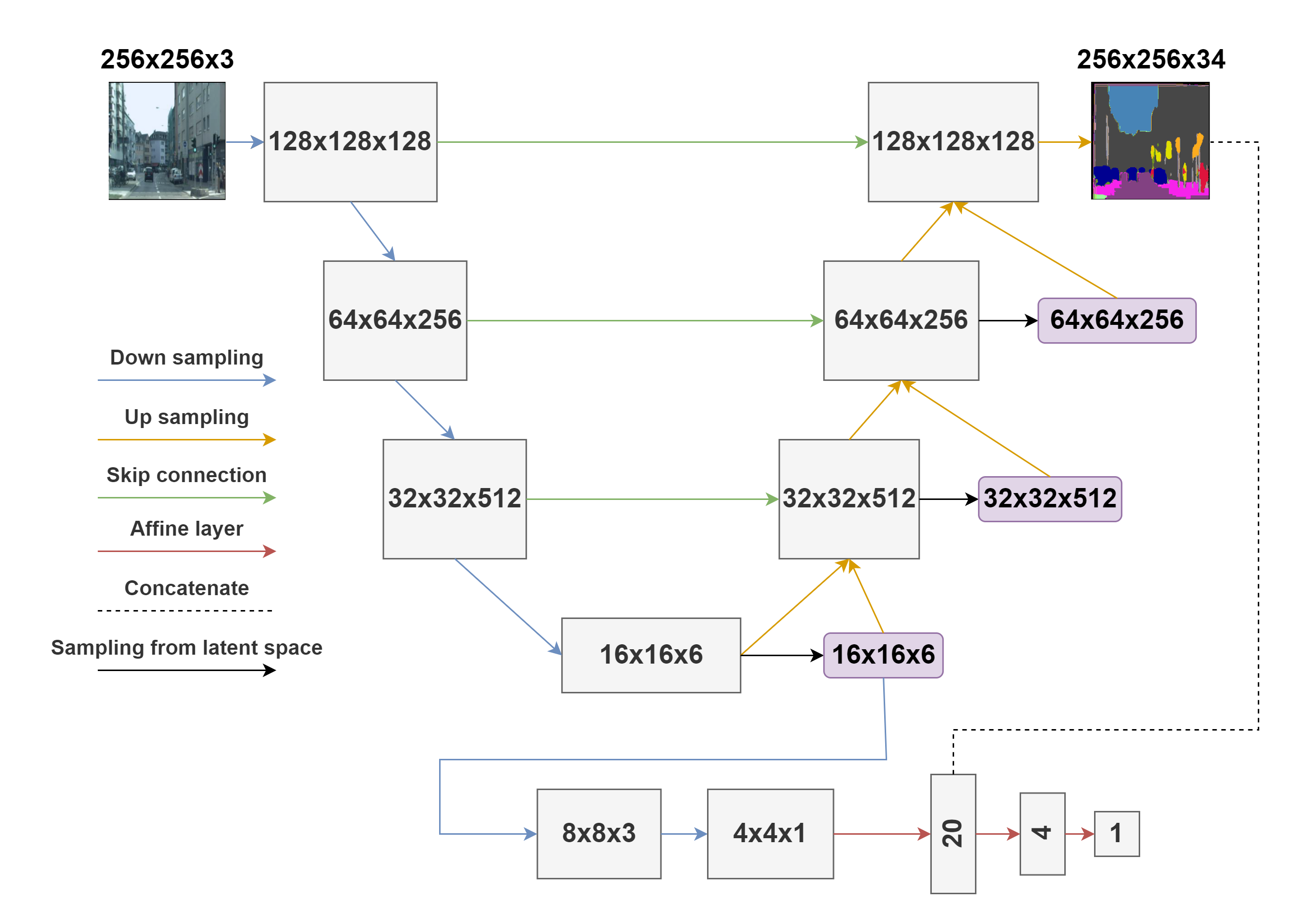}
        \caption{Latent friction architecture}
        \label{fig:latentFriArch}
\end{figure}

\begin{figure}[!h]
     \centering
         \includegraphics[width=0.5\textwidth]{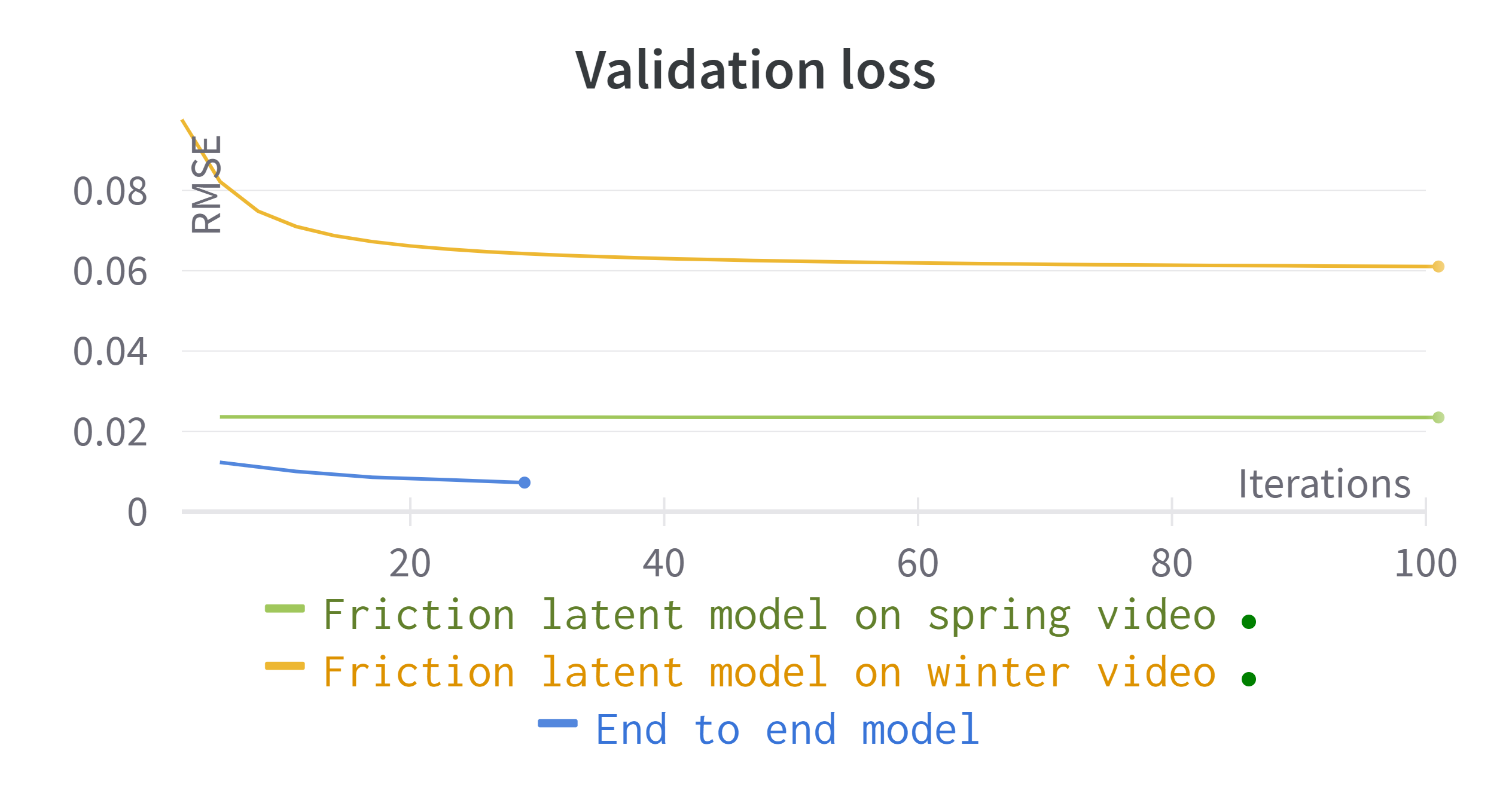}
        \caption{Compare of friction estimation models}
        \label{fig:CompValLossFri}
\end{figure}

As explained above, in the second experiment, the segmentation model is utilized as feature extractor. The overall architecture is shown in Fig \ref{fig:latentFriArch}. For this experiment, the UNet architecture is frozen and the friction estimator part is trained. Using this architecture, two models are trained on two different videos. The first video is captured in May and only contains sunny days of spring and the other video is captured in February when everywhere covered with snow. The purpose of doing this is to investigate whether the segmentation model trained on Cityscapes dataset, which does not include any winter condition, can extract useful information from the wintery scenes and whether the extracted features can be transferred. The RMSE of these experiments are compared in Fig \ref{fig:CompValLossFri}.

As Fig \ref{fig:CompValLossFri} shows, the segmentation model could extract some useful information even from driving scenes in wintery days. However, as expected the segmentation model performs better on footage captured in spring, since the model is trained on images from sunny days. It may not be obvious in Fig \ref{fig:CompValLossFri} but the loss value on the spring video slightly decreases. The end-to-end model outperforms the other two models, although the friction latent model on the spring video has come very close to the performance of the end-to-end model. A summary of the models' best RMSEs is shown in Table \ref{Tab:CompValLossFri}.

\begin{table}[]
\centering
\begin{tabular}{|c|c|}
\hline
\textbf{Model name}      & \textbf{RMSE} \\ \hline
Friction latent - spring & 0.04853       \\ \hline
Friction latent - winter & 0.02348       \\ \hline
End-to-end model         & 0.007267      \\ \hline
\end{tabular}
\caption{Compare of friction estimation models RMSE}
\label{Tab:CompValLossFri}
\end{table}

% \section{Discussion}

% ...

\section{Conclusions}

Aiming to estimate frictional force, prior works either used expensive sensors like \ac{nir} cameras or proposed methods that are not practical. Some others rely on developing dynamical models for real-time responses. Using deep semantic segmentation and regression method, a new method is proposed in this work to estimate friction. Inspired by the way drivers visually guess the friction of the surface ahead, the environmental elements are scrutinized by semantic segmentation and the result is utilized to estimate frictional force.

As there is no visual dataset available to estimate frictional force, thanks to the Volvo Trucks team, we were given access to Volvo testing centers and test trucks to create a visual vehicle dynamic dataset. This dataset was used to train a deep regression model to estimate frictional force. In addition, using the Cityscapes dataset, a semantic segmentation model was trained. Three experiments have been carried out to show whether the idea of using joint semantic segmentation and regression can be used for frictional force estimation. While the outcomes look promising there is still a lot of scope to improve the preliminary results.

As the Cityscapes dataset only contains one surface category, future work will aim to create a dataset where a number of desired surface types are included. Thus, different weather conditions and surface types can be covered. Moreover, the idea of using a semantic segmentation latent space to estimate friction requires further exploration as only preliminary results are presented here.  

\begin{table}[]
\centering
\begin{tabular}{|c|c|}
\hline
\textbf{Method}                                                                      & \textbf{IoU} \\ \hline
Ours                                                                                 & 0.69         \\ \hline
\begin{tabular}[c]{@{}c@{}}A Hierarchical Probabilistic\\ U-Net {[}4{]}\end{tabular} & 0.62         \\ \hline
\end{tabular}
\end{table}

\section*{Acknowledgment}

The authors would like to thank Volvo Trucks for providing financial support of this project. Volvo Trucks also gave us a unique opportunity to use their facilities to collect valuable data without which this research was infeasible. 

\bibliographystyle{IEEEtran}
\bibliography{bib}

\end{document}